# IMPROVING LOCAL SEARCH FOR FUZZY SCHEDULING PROBLEMS

M J Geiger, S Petrovic
Automated Scheduling, Optimisation & Planning Group, School of Computer Science & IT, University of Nottingham

**Key words to describe the work:** Fuzzy scheduling, dynamic scheduling, multi criteria optimization, local search.

**Key Results:** In the investigated problem, multi criteria local search strategies may outperform their single criterion counterparts.

**How does the work advance the state-of-the-art?:** The proposed technique successfully improves current local search heuristics for fuzzy scheduling problems that bear fitness plateaus in the search space.

**Motivation (problems addressed):** Although there do exist various approaches for real-world scheduling problems, the methodology is not yet fully investigated and understood. In particular the field of fuzzy scheduling is a new aspect of optimization in real-world problems.

## Introduction

The integration of fuzzy set theory and fuzzy logic into scheduling is a rather new aspect with growing importance for manufacturing applications [1, 4], resulting in various unsolved aspects.

In the current paper, we investigate an improved local search technique for fuzzy scheduling problems with fitness plateaus, using a multi criteria formulation of the problem. We especially address the problem of changing job priorities over time as studied at the Sherwood Press Ltd, a Nottingham based printing company, who is a collaborator on the project.

## A Fuzzy Scheduling Problem in the Printing Industry

In the studied problem, a set of machines $M = \{M_1,...,M_m\}$ exists performing operations on a set of jobs $J = \{J_1,...,J_n\}$. Each job $J_i$ can be described as an ordered set of tasks $T_i = \{T_{i1},...,T_{it}\}$ whereas each task may be processed on a specific machine with a nonnegative processing time $p_{ij}$.

The relative importance of the jobs is modelled by introducing weights $w_i$ for each job $J_i$ which may change over time.

As studied at Sherwood Press Ltd, due dates may exist as so called 'promised delivery dates', based upon an individual agreement with each customer. Here, the products are supposed to be shipped within a time interval and not necessarily on a specific day. Accordingly, the notion of tardiness of a job $J_i$ can properly be addresses using a fuzzy due date definition $\widetilde{D}_i$, consisting of a doublet $(d_i^1, d_i^2)$ with monotonically decreasing satisfaction degree and a correspondingly increasing fuzzy tardiness $\widetilde{T}_i$ as shown in fig. 1.

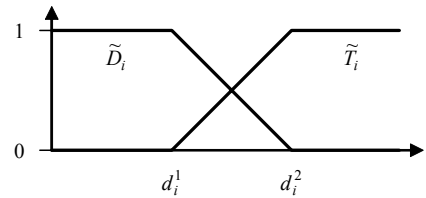

Fig 1. Fuzzy due date $\widetilde{D}_i$ and fuzzy tardiness definition $\widetilde{T}_i$.

In addition a tardiness penalty $T_i'$ has been introduced, allowing penalty values > 1, as described in fig. 2.

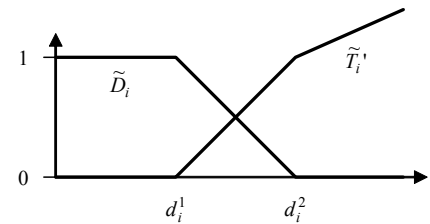

Fig. 2. Tardiness penalty $T_i'$.

Two overall objective definitions have been studied. First, the single-criterion minimization of the total weighted tardiness $c_1(x) = \sum w_i \widetilde{T}_i$, and second the bicriteria extension as a vector optimization problem $\min(c_1, c_2)$ with $c_2(x) = \sum w_i T_i'$, leading to the problem of identifying all Pareto optimal solutions [2].

The introduction of the tardiness penalty as described in fig. 2 is motivated by the existence of fitness plateaus in the search space with respect to

$c_1$, as there do exist numerous alternatives $x$ with identical objective values $c_1$.

**Investigation of Local Search Strategies**
Three basic local search neighbourhoods have been investigated within a hillclimbing framework [3].
- Forward shift (FSH), removing a job from a current schedule and reinserting it after a succeeding one.
- Backward shift (BSH), removing a job from its position and reinserting it before a preceding job.
- Exchange (EX), exchanging the position of two jobs.

Each neighbourhood structure has been allowed to compute 20,000 alternatives while the weights $w_i$ have been altered after 5,000, 10,000 and 15,000 iterations in order to simulate dynamic changing job priorities.

The tests have been performed on 50 problem instances generated based on the job characteristics of the practical case in the Sherwood Press, and the results have been averaged.

**Results**
Conducting significance tests at a level of 0.99, it has been possible to observe that for the single criterion formulation EX leads in 90% of the instances to better results than BSH, which is found to be better than FSH. The results are stable over time, thus the altering of the weights does not lead to significantly different results.

The comparison of the neighbourhood search on the bicriteria extension of the problem with the results of the single criterion version reveals an interesting pattern, shown in fig. 3. After approximately 1,000 initial iterations, the single criterion search is able to significantly outperform the bicriteria formulation in more problem instances than vice versa. However, this effect is reversed after the local search approaches have been allowed more evaluations. While the monocriterion hillclimber tends to get stuck in local optima, its bicriterion counterpart successfully overcomes local optimality, leading to significantly better results.

As fig. 3 reveals, this effect is each time repeated after changing the weight settings. While the neighbourhood search for the single criterion problem allows comparably fast improvements, its advantage over the multi criterion extension is decreasing over time, and the results are reversed given the necessary amount of iterations.

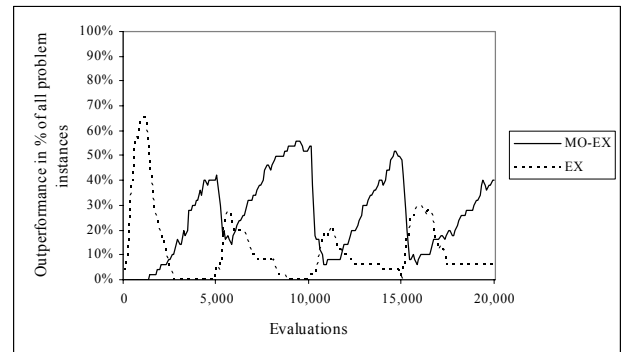

Fig 3. Comparison of the bicriteria MO-EX with the EX-neighbourhood. The results of a statistical significance test have been aggregated over all problem instances.

**Conclusions**
A fuzzy scheduling problem derived from a practical application in the printing industry has been presented. It is possible to show that for the studied problem, multi criteria local search strategies may outperform their single criterion counterparts at the cost of additional computations.

**Acknowledgements**

We would like to thank our collaborators at Sherwood Press Ltd, Nottingham, Mr Colin Edis, Mr James Timpson and Ms Jenny Bull.
This research is supported by the British Engineering and Physical Sciences Research Council (EPSRC), UK (Grant No. GR/R95319/01).